\documentclass[final,1p,times]{elsarticle}
\usepackage{graphicx}

\usepackage{amssymb}
\usepackage{amsmath}
\usepackage{microtype}
\usepackage{caption}
\usepackage{subcaption}
\usepackage[acronym, style=super, section=section, nonumberlist, nomain,
nopostdot]{glossaries}
\usepackage[]{glossaries-extra}
\makeglossaries

\setabbreviationstyle[acronym]{long-short}
\newacronym[longplural={Debugging Information Entities}]{DIE}{DIE}{Debugging Information Entity}
\newacronym[shortplural={OSes}, firstplural={operating systems (OSes)}]{OS}{OS}{operating system}

\newacronym{IQL}{IQL}{Independent Q‑Learning}

\newacronym[first={NVIDIA OpenSHMEM Library (NVSHMEM\texttrademark)}]{NVSHMEM}{NVSHMEM}{NVIDIA OpenSHMEM Library}
\newacronym[shortplural={ANNs}, firstplural={Artificial Neural Networks}]{ANN}{ANN}{Artificial Neural Network}
\newacronym[shortplural={CNNs}, firstplural={Convolutional Neural Networks}]{CNN}{CNN}{Convolutional Neural Network}
\newacronym[shortplural={VAEs}, firstplural={Variational Autoencoders}]{VAE}{VAE}{Variational Autoencoder}
\newacronym[shortplural={GANs}, firstplural={Generative Adversarial Networks}]{GAN}{GAN}{Generative Adversarial Network}

\newacronym{KTH}{KTH}{KTH Royal Institute of Technology}
\newacronym{SAR}{SAR}{Synthetic Aperture Radar}
\newacronym{LSWI}{LSWI}{Land Surface Water Index}
\newacronym{NDWI}{NDWI}{Normalized Difference Water Index}
\newacronym{DEM}{DEM}{Digital Elevation Model}
\newacronym{NIR}{NIR}{Near Infra-Red}
\newacronym{NDVI}{NDVI}{Normalized Difference Vegetation Index}
\newacronym{RGB}{RGB}{Red Green Blue}
\newacronym{LSTM}{LSTM}{Long Short-Term Memory}
\newacronym{IOU}{IoU}{Intersection over Union}

\newacronym{LAN}{LAN}{Local Area Network}
\newacronym{VM}{VM}{virtual machine}
\newacronym{WiFi}{Wi‑Fi}{Wireless Fidelity}

\newacronym{WLAN}{WLAN}{Wireless Local Area Network}
\newacronym{UN}{UN}{United Nations}
\newacronym{SDG}{SDG}{Sustainable Development Goal}

\usepackage{lineno}

\journal{International Journal of Applied Earth Observation and Geoinformation}

\begin{document}

\begin{frontmatter}



\title{AquaCluster: Using Satellite Images And Self-supervised Machine Learning Networks To Detect Water Hidden Under Vegetation}


\author[label1]{Ioannis Iakovidis}\ead{iiak@kth.se} 
\author[label1]{Zahra Kalantari}\ead{zahrak@kth.se} 
\author[label2]{Amir H. Payberah}\ead{payberah@kth.se} 
\author[label3]{Fernando Jaramillo}\ead{fernando.jaramillo@natgeo.su.se} 
\author[label2]{Francisco J. Peña\corref{cor1}}\ead{frape@kth.se}\cortext[cor1]{Corresponding author} 

\affiliation[label1]{organization={Division of Water and Environmental Engineering, KTH Royal Institute of Technology},
            addressline={Teknikringen 10b}, 
            city={Stockholm},
            postcode={11428}, 
            country={Sweden}}

\affiliation[label2]{organization={Division of Software and Computer Systems, KTH Royal Institute of Technology},
            addressline={Kistagången 16}, 
            city={Kista},
            postcode={164 40}, 
            country={Sweden}}
\affiliation[label3]{organization={Department of Physical Geography, Stockholm University},
            addressline={Svante Arrhenius väg 8}, 
            city={Stockholm},
            postcode={10691}, 
            country={Sweden}}
\begin{abstract}
In recent years, the wide availability of high-resolution radar satellite images has enabled the remote monitoring of wetland surface areas. Machine learning models have achieved state-of-the-art results in segmenting wetlands from satellite images. However, these models require large amounts of manually annotated satellite images, which are slow and expensive to produce. The need for annotated training data makes it difficult to adapt these models to changes such as different climates or sensors. To address this issue, we employed self-supervised training methods to develop a model, AquaCluster, which segments radar satellite images into water and land areas without manual annotations. Our final model outperformed other radar-based water detection techniques that do not require annotated data in our test dataset, having achieved a 0.08 improvement in the Intersection over Union metric. Our results demonstrate that it is possible to train machine learning models to detect vegetated water from radar images without the use of annotated data, which can make the retraining of these models to account for changes much easier.
\end{abstract}




\end{frontmatter}



\section{Introduction}
\label{sec:intro}
Wetlands cover only a small part of the Earth’s surface, but they play a crucial role in supporting the environment and protecting against climate impacts. 
They help clean water, reduce the risk of floods, and store large amounts of carbon, making them essential for sustainable development~\cite{jaramillo2019priorities}. 
However, wetlands are being lost at an alarming rate due to climate change and human activities such as agriculture and urban development~\cite{thorslund2017wetlands}. 
A major challenge in protecting wetlands is accurately tracking their surface water, especially when it is hidden beneath vegetation. 
This highlights the need for better monitoring tools and techniques. 

High-resolution satellite imagery has become widely available in recent years, making it easier to monitor water bodies remotely across the globe. A common method involves using optical satellite images to estimate the surface area of water. This works well for open water, where detection is relatively straightforward~\cite{SatelliteDetectionofSurfaceWaterExtent}. However, wetlands and other water bodies that are partly or fully covered by vegetation are more difficult to detect. Optical sensors cannot see through vegetation, so they often miss water that is hidden beneath it. In contrast, radar sensors can penetrate vegetation and detect water underneath~\cite{ozesmi2002satellite}. But radar images often contain noise, such as speckle, which makes it harder to distinguish water from land~\cite{speckle_noise}. Given these limitations, data-driven approaches such as machine learning have become increasingly attractive for classifying complex environments like wetlands.

Classifying water in radar images, especially in vegetated wetlands, is particularly difficult using traditional machine learning methods. Previous papers have used methods such as decision trees~\cite{townsend2002relationships}, random forests~\cite{rs11050593} and Bayesian probability thresholding~\cite{8127439} to detect open and vegetated water. Many researchers have also taken advantage of the availability of multitemporal radar satellite images to apply machine learning methods to time series of radar data for the same task~\cite{rs11060720, 10.1117/1.JRS.19.021006}. Another common approach is to combine optical and radar data to exploit the advantages of each~\cite{SLAGTER2020102009, rs15143495}.

In recent years, researchers have increasingly turned to deep learning models to tackle image tasks, due to their ability to automatically learn spatial and spectral patterns from image data. Among these, \glspl{CNN}~\cite{gu2018recent} have performed strongly, particularly in semantic segmentation tasks, where each pixel of the input image is assigned a class label~\cite{HAO2020302}. In the last decade \glspl{CNN} have been established as the most widely used machine learning method for the task of semantic segmentation~\cite{guo2018review}. Unlike traditional classification approaches that label entire image regions, semantic segmentation provides the fine-grained detail needed to map heterogeneous landscapes like wetlands. This level of resolution is essential for detecting small, fragmented, or vegetation-covered water bodies that might otherwise go unnoticed. Specifically for the task of wetland detection, \glspl{CNN} that combine information from optical and radar images have been applied~\cite{9803207, rs12010002, 9576524}, outperforming simpler methods. 

Although \gls{CNN}-based models perform well on semantic segmentation tasks, they typically rely on large amounts of labeled data for training. In remote sensing, producing such annotated datasets is particularly challenging due to the cost and effort involved in manually labeling satellite images. This means that most \gls{CNN}-based models trained in wetland detection rely  on global, national or regional datasets like the CORINE Land Cover inventory~\cite{cover2018copernicus} or the Climate Change Initiative Land Cover project~\cite{cci2017european} that are updated infrequently and do not offer the temporal resolution necessary to accurately map seasonal water areas like wetlands.

This reliance on supervision not only hinders the performance of the machine learning models but also limits their scalability and broader application. It is often hard to retrain those models to adapt them to different conditions, such as other climates or sensors, due to lack of relevant annotated data. To address problems like this, researchers have developed alternative training strategies that reduce or eliminate the need for labeled data. One promising approach is self-supervised learning, which trains models using tasks that are automatically generated from the input data itself~\cite{rani2023self}.

Self-supervised learning methods have recently performed strongly in remote sensing applications, including land cover classification and semantic segmentation~\cite{Self-supervisedLearningRemoteSensing}. In the context of wetland mapping, a self-supervised model that combines optical and radar data~\cite{pena2023deepaqua} has been developed and used to track water surface changes in multiple sets of wetlands in Scandinavia~\cite{egusphere-2024-3248}. However, this model relies on the complementary strengths of both modalities and is not applicable in radar-only settings. In many regions, persistent cloud cover or dense vegetation limits the availability of optical data, making radar the only reliable option for consistent wetland monitoring.

To address the above limitations, we developed AquaCluster, a \gls{CNN}-based self-supervised model that segments wetlands into water and non-water areas using only radar satellite images. Our approach builds on the framework introduced in~\cite{Unsupervised}, which combines two complementary self-supervised techniques to improve the model’s ability to separate land and water in noisy radar data. To ensure accessibility and efficiency, we used lightweight model architectures and implemented an ensemble version of the model~\cite{sagi2018ensemble} by combining the predictions of multiple models to further boost segmentation accuracy.

In summary, this paper makes the following key contributions:
\begin{itemize}
    \item We propose AquaCluster, a fully self-supervised machine learning framework for the semantic segmentation of wetlands using only radar satellite imagery, addressing the challenge of detecting water beneath vegetation without relying on labeled data.
    \item To enhance both accuracy and stability, we introduce an ensemble version of the model that combines predictions from multiple, independently trained networks.
    \item We demonstrate that our ensemble AquaCluster model outperforms the baseline statistical method Otsu on the same dataset, as well as the optical-based dynamic world model.
\end{itemize}

The paper is structured as follows. In Section \ref{sec:related_work}, we present other work related to our own. In Section \ref{sec:background}, we give a brief overview of the field. In Section \ref{sec:method}, we provide a detailed description of our methods. In Section \ref{sec:experiments}, we explain our experiment setup and, in Section \ref{sec:results}, we present the results of our experiments. Finally, in Section \ref{sec:conclusion}, we present our conclusions.

\section{Related Work}
\label{sec:related_work}

In this section we present related works that seek to apply machine learning methods to the task of wetland detection and works that concern self-supervised machine learning methods applied to remote sensing tasks.

\subsection{Applying machine learning in wetland detection} Early attempts at segmenting remote sensing images using machine learning methods consisted of applying simple machine learning classification methods, such as random forests and support vector machines, to individual pixels~\cite{LARY20163}. Mahdianpari et. al. were the first to test \gls{CNN}'s image processing capabilities on this task~\cite{mahdianpari2018very} and tested a wide variety of \gls{CNN} architectures. The paper concluded that most of the \glspl{CNN} tested achieved state-of-the-art results on the task of wetland mapping, outperforming simpler machine learning methods such as random forests. Moreover, the study found that pretraining the \gls{CNN} on a generic computer vision optical image dataset did not improve performance. This is due to the fact that models pretrained on optical image datasets (which most machine learning image training datasets are) can only make use of the red, green and blue spectral bands of satellite images. Moreover, satellite images of landscapes differ significantly from the images found in most computer vision training datasets. Since then, several papers have confirmed the high performance of \gls{CNN}-based models on wetland detection and classification, especially when utilizing a variety of different data types such as optical images, radar images, digital elevation models or images from different seasons~\cite{mahdianpari2018very, jiang2019arcticnet, Labrador, onojeghuo2023wetlands}.

In order to further improve performance, other authors have employed more complex architectures. Pham et al.~\cite{pham2022new} utilized a \gls{CNN} with two paths, a shallower one to extract spatial information and a deeper one to extract more high-level context information from the input images. The two paths are joined using an attention mechanism~\cite{vaswani2017attention}, which enables the model to put greater focus on the more important part of the information contained in each of the model's layers. Cui et. al.~\cite{cui2020wetlandnet} used a modified U-Net architecture~\cite{ronneberger2015u} to improve the semantic segmentation of coastal wetlands in China by processing each spectral channel individually before combining all the channel information together. This method drastically reduces the number of trainable parameters in the model, which in turn decreases the data needed for training.

\subsection{Applying self-supervised machine learning in remote sensing} The fact that there are multiple satellites with different sensors providing observation data on various dates means that there are many different images available for any geographical area. Many self-supervised training methods train machine learning models to produce similar outputs for pairs of different images of the same area. These are called positive pairs. At the same time, these methods train models to produce dissimilar outputs for pairs of images of different areas, called negative pairs. These methods belong in the category of contrastive learning~\cite{jaiswal2020survey}. Scheibenreif et. al.~\cite{scheibenreif2022contrastive} adapted the SimCLR contrastive learning method~\cite{chen2020simple} to perform multi-label image classification of satellite images by using two different models, one for optical images and one for radar images. Contrastive learning is achieved by using images of the same area but from different sensors as positive pairs, while images from different areas are used as negative pairs. The authors found that using self-supervised learning and then fine-tuning the resulting model with 10\% of the annotated training data outperformed a fully-supervised model. The authors used a similar method in the paper "Self-supervised vision transformers for land-cover segmentation and classification"~\cite{scheibenreif2022self}. As with the previous paper, optical and radar images of the same areas were fed into two models and the data extracted by the two images were compared using contrastive learning. This paper used \glspl{CNN} enhanced with an attention mechanism as encoders and again found that pretraining with self-supervision and using 10\% of the training dataset outperformed fully-supervised training methods.

Manas et. al.~\cite{manas2021seasonal} took advantage of the multitemporal nature of satellite data by using satellite images of the same area during different seasons to create positive pairs for contrastive learning. This seasonal augmentation is complemented with other traditional augmentations to produce positive groups of images from multiple augmentation methods. As usual, images of other areas (augmented and not) are used as negative samples. The authors found that this self-supervision method outperforms pretraining on a computer vision optical image dataset and Momentum Contrast pretraining~\cite{he2020momentum} in the change detection task in satellite images.

Another method that can be used to generate similar and dissimilar pairs of data points for contrastive learning is unsupervised clustering methods. For each data point the model produces a vector of values as a representation of that data point. Then a clustering method such as K-means~\cite{ahmed2020k} is used to separate these representations into groups. Cai et. al.~\cite{cai2021task} used an unsupervised image segmentation algorithm to segment remote sensing images and used tiles from the same segment as positive pairs and tiles from different segments as negative pairs to pretrain a \gls{CNN} model for change detection.

\section{Background}
\label{sec:background}

\subsection{Remote monitoring of wetlands}
Wetlands are land areas whose soil is either permanently or seasonally inundated with water. They are important hubs of biodiversity~\cite{Dataforwetlandscapes} and store enormous amounts of greenhouse gases~\cite{leifeld_underappreciated_2018}. Unfortunately, due to land use change and global warming wetlands around the world are rapidly disappearing. This has resulted in the eradication of over half of all wetlands globally since the start of the twentieth century~\cite{davidson2014much}. Therefore the continuous monitoring of wetlands is necessary.

Although in-situ monitoring of wetlands is the most precise method~\cite{rs70810938}, the large number of wetlands all over the world means it is very impractical to monitor them all this way. In recent years the availability of high-resolution satellite images of the Earth's surface has greatly increased. These satellite images have allowed the remote monitoring of wetlands worldwide~\cite{ozesmi2002satellite}. One of the ways satellite images can be used to monitor wetlands is to extract the location and water surface area of wetlands.

\begin{figure}[t]
  \centering
  \includegraphics[width=0.8\linewidth]{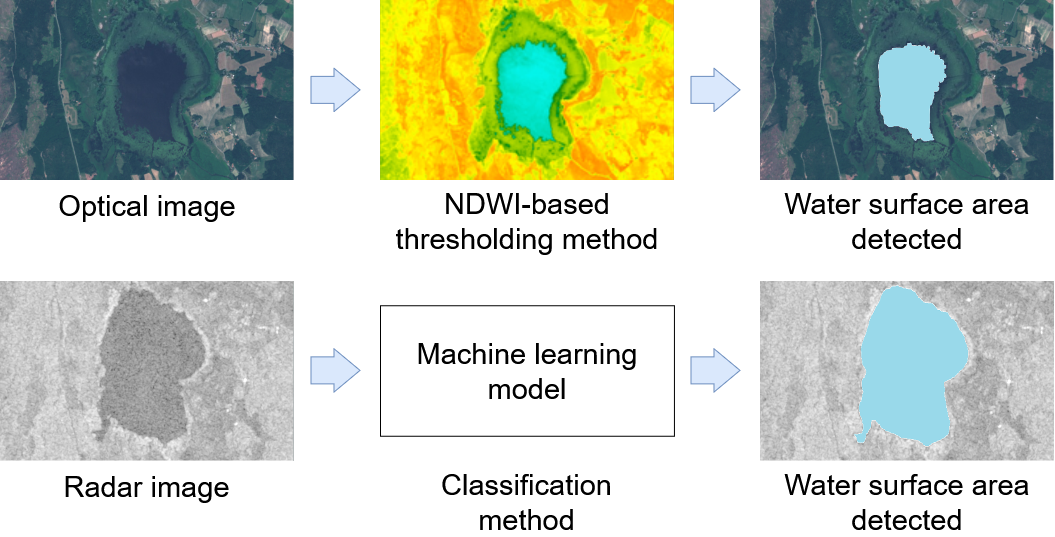}
   \caption{Comparison of methods using optical and radar satellite images for water detection used on a wetland partially covered by vegetation. We can see that the method based on optical data cannot detect the water hidden under vegetation. On the other hand the radar image contains a lot of noise which necessitates a more complicated classification method. Modified from \cite{pena2023deepaqua}.}
   \label{fig:optical-radar}
\end{figure}

In order to detect bodies of open water from remote sensing images, optical satellite images have traditionally been used. The method most often used is the Normalized Difference Water Index (NDWI) method, which takes advantage of the different reflectance properties of water compared to vegetation and soil~\cite{ndwiref}. By comparing the amount of green and near-infrared light reflected from an area, we can separate open water from land in most circumstances.

Unfortunately, wetlands are often partly or fully covered by vegetation, which optical signals cannot penetrate. This problem can be solved by using radar instead of optical satellite images, since radar signals such as the Sentinel-1 C-band sensor can penetrate vegetation~\cite{ozesmi2002satellite}. Another advantage of radar signals is their ability to penetrate clouds, allowing the monitoring of water bodies regardless of any cloud coverage. Unfortunately, radar images are affected by speckle noise, which makes the task of separating water from land non-trivial~\cite{speckle_noise}. Recently various machine learning models have been applied to this task. A comparison of the use of optical and radar satellite images for monitoring the water extent of wetlands can be seen in Figure \ref{fig:optical-radar}.

\subsection{Applying machine learning models to satellite imagery}\label{sec:back-2}
\glspl{CNN}~\cite{gu2018recent} are a category of machine learning models designed explicitly to operate on images and have achieved great success in the field of image processing during the last few years. \glspl{CNN} have achieved state of the art results in the semantic segmentation of remote sensing images~\cite{YUAN2021114417}, such as detecting hydrological barriers in wetlands~\cite{HUBINGER2024114314}. However, the huge number of annotated data \glspl{CNN} require for training severely limits their applicability in many areas. Many modern \glspl{CNN} designed for image-based tasks are trained using millions of annotated images~\cite{5206848}. While there are a number of image datasets designed for training machine learning models, the vast majority of them consist of optical images of everyday objects and scenes. These are, therefore, not suitable for training models that take remote sensing data as input. Creating new image training datasets is time-consuming and expensive, especially when talking about semantic segmentation datasets, which require  every pixel in each image to be annotated. The problem is even worse in the field of remote sensing, where annotating data such as satellite images often requires expert knowledge or even in-situ measurements.

In order to overcome these limitations of machine learning models, a group of training methods, called self-supervised training methods, have been developed. These methods rely on training the machine learning model on tasks that can be automatically generated from the input data~\cite{rani2023self}.

Self-supervised learning methods include contrastive training, which focuses on comparing the model encodings of multiple data points~\cite{jaiswal2020survey}. More specifically, contrastive training methods create groups of semantically similar data points and train the machine learning model to output similar encodings for them. In order to prevent the model from just outputting the same encoding for any input, groups of semantically dissimilar data points are created. The model is trained to output dissimilar encodings for points in those groups.

One sub-category of contrastive training is deep clustering~\cite{caron2018deep}. Deep clustering training methods start by passing the training dataset through the untrained model, which outputs an encoding for each data point in the dataset. These encodings are then clustered using a clustering algorithm. The cluster assigned to each data point is treated as its label and these pseudo-labels are used to update the model weights. This process of obtaining data encodings via the model, clustering these encodings, and using the cluster assignments in place of manual labels to update the model weights is repeated until the model has been fully trained. Via this process the model is trained in a classification task without needing any manual annotations.

Another sub-category of contrastive training is negative sampling~\cite{pmlr-v119-chen20j}. Negative sampling algorithms compare the encodings of pairs of data points. The pairs of semantically similar data points are called positive pairs. They usually consist of a data point from the training dataset and a slightly modified (also called augmented) version of it. The pairs of semantically dissimilar data points are called negative pairs and they are usually two random data points from the training dataset. These algorithms often utilize a Siamese model architecture~\cite{chicco2021siamese} with two parallel sub-models, the first one for processing the original data and the second one for processing the augmented data. After training, the second model is discarded and the first model is used for inference.

It should be noted that when using fully self-supervised training methods without any annotated data we can only control the number of classes the model will divide the data into, but not what each of those classes represents. For this reason self-supervised segmentation algorithms tend to perform better when asked to segment the input image into more classes than the number of ground truth classes~\cite{Unsupervised} (which, in our case, are water and land). Consequently, for self-supervised classification models, some amount of post-processing is required to correspond the class assignments that the model outputs to the ground truth classes of the data.

\subsection{Ensemble machine learning models}

Ensemble learning~\cite{sagi2018ensemble} refers to a field of machine learning that combines the predictions of multiple models to achieve improved performance and reduced variance over a single model. Most common ensemble machine learning methods achieve this by training a number of models of identical architecture with different subsections of the training data and then combining their predictions on test data using an (often weighted) voting method. Self-supervised models have also been found to benefit from ensembling~\cite{ruan2023weightedensembleselfsupervisedlearning}.

\section{Method}
\label{sec:method}

In this section, we describe the training and inference algorithms of the AquaCluster model, as well as the architecture of the sub-models used.

\subsection{Training}
\begin{figure}[!hb]
\centering
\includegraphics[scale=0.245]{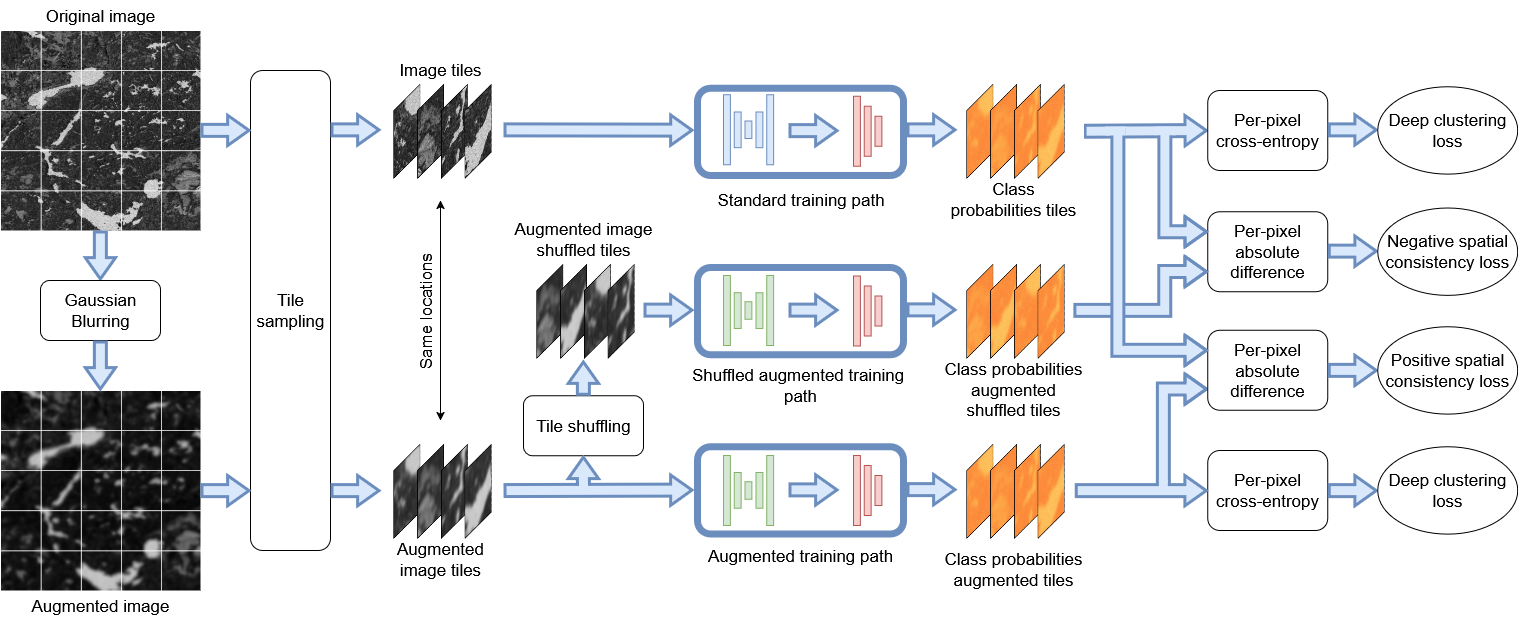} 
\caption{The training algorithm. Three groups of tiles are sampled from the training image (image tiles, augmented image tiles and augmented image shuffled tiles). Each of these groups of tiles is processed by an encoding sub-model and the prediction sub-model to produce tiles of pixel-level class probabilities. Using the class probability tiles and two different loss algorithms (per-pixel cross-entropy and per-pixel absolute difference), the four losses used to train the sub-models (two deep clustering losses, a positive spatial consistency loss and a negative spatial consistency loss) are computed. Sub-models with the same color share parameters.}
\label{fig:training-algorithm}
\end{figure}

In order to train a machine learning model for the task of the semantic segmentation of radar satellite images to detect wetland water surfaces, we used a deep clustering training algorithm, as it is one of the most studied categories of self-supervised learning with a large number of applications in the field of remote sensing~\cite{Self-supervisedLearningRemoteSensing}. Inspired by~\cite{Unsupervised}, we enhanced the deep clustering training by incorporating negative sampling, taking advantage of the spatial information contained in satellite images. During training the AquaCluster model processes the input images and produces a representation (also called an encoding) for each pixel. In order to apply a deep clustering loss to the model, these encodings need to be assigned pseudo-labels. Instead of a traditional clustering algorithm, we employ a small untrained \gls{CNN} sub-model, which we expect to assign similar pseudo-labels to input pixels with similar embeddings even before training. These pseudo-labels are used to compute the deep clustering loss. In order to apply a negative sampling loss to the model we need to compare the class assignments of each pixel with those of a similar and a dissimilar pixel. To generate similar (or positive) pixel pairs we apply Gaussian blurring to the input images and consider each blurred pixel a positive pair of the corresponding pixel of the original input image. To generate dissimilar (or negative) pixel pairs we consider each blurred pixel a negative pair of a random pixel of the original input image. We then apply a negative sampling loss by training the model to output similar class probabilities for positive pairs of pixels and dissimilar probabilities for negative pairs of pixels. An illustration of the AquaCluster training algorithm can be found in Figure \ref{fig:training-algorithm}.

The training algorithm steps of AquaCluster are:

\begin{enumerate}
    \item The AquaCluster training data, which are radar satellite images of a large geographical area, are split into square \textit{image tiles}. 
    \item The image tiles produced in step $1$ are processed in the \textit{standard training path} to produce tiles of per-pixel class probabilities.
    \item For each pixel in the tiles of class probabilities produced in step $2$, the class with the highest probability is treated as the pixel's label, and a cross-entropy loss is calculated. The sum of these losses over each pixel of each tile is the \textit{first deep clustering loss} $L_c$.
    \item The training data is augmented with a Gaussian blur filter~\cite{373563} before being split into square \textit{augmented image tiles}. The ordering of the splitting algorithm employed in steps $1$ and $4$ is consistent, meaning that any augmented image tile produced in this step depicts the same geographical area as the corresponding image tile produced in step $1$.
    \item The augmented image tiles produced in step $4$ are processed in the \textit{augmented training path} to produce augmented tiles of per-pixel class probabilities.
    \item For each pixel in the augmented tiles of class probabilities produced in step $5$, the class with the highest probability is treated as the pixel's label, and a cross-entropy loss is calculated. The sum of these losses over each pixel of each tile is the \textit{second deep clustering loss} $\hat{L}_c$.
    \item The augmented image tiles produced in step $4$ are shuffled to produce \textit{augmented image shuffled tiles}. The shuffling means that any tile produced in this step depicts a different geographical area than the corresponding image tile produced in step $1$.
    \item The augmented image shuffled tiles produced in step $7$ are processed in the \textit{augmented shuffled training path} to produce augmented shuffled tiles of per-pixel class probabilities.
    \item Each image tile forms a positive pair with the corresponding augmented image tile, since they depict the same area. To train AquaCluster to produce similar predictions for positive pairs, we use the sum of the per-pixel absolute difference in class probabilities over each pixel pair as the \textit{positive spatial consistency loss} $L_p$.
    \item Each image tile forms a negative pair with the corresponding augmented image shuffled tile, since they depict different areas. To train AquaCluster to produce dissimilar predictions for negative pairs, we use the negative sum of the per-pixel absolute difference in class probabilities over each pixel pair as the \textit{negative spatial consistency loss}  $L_n$.
    \item The weighted sum of the four losses $L_c$, $\hat{L}_c$, $L_p$ and $L_n$ is the \textit{total loss of the training algorithm}, denoted as $L$.
\end{enumerate}
In the following subsections we will provide more detail on the training paths, the losses of the algorithm, and the architecture of the sub-models.

\begin{figure}[!hb]
\centering
\includegraphics[scale=0.35]{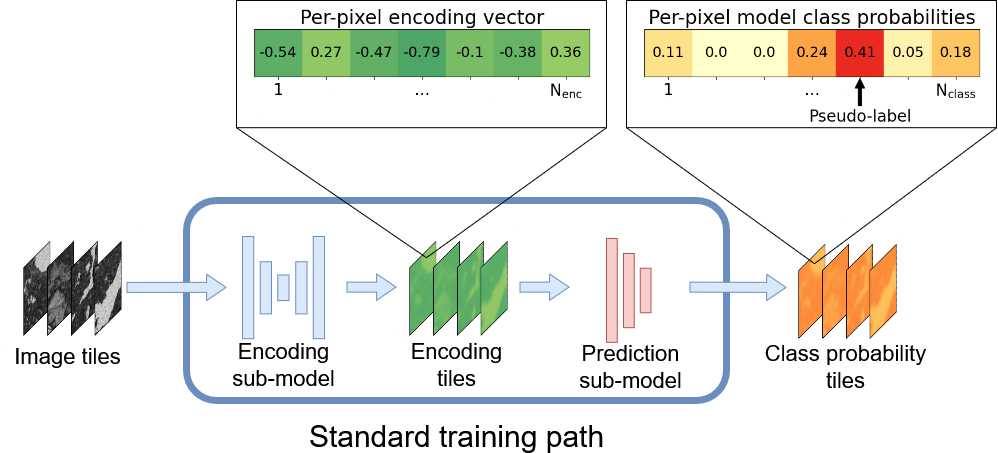} 
\caption{The standard path of the training algorithm. Each image tile is processed by the encoding sub-model, which outputs tiles containing an $N_{enc}$-sized encoding vector for each pixel. These tiles are then processed by the prediction sub-model, which outputs tiles with class probabilities for each pixel. For each pixel the class with the highest probability is treated as its label, which is used in the deep clustering loss. The class probabilities of each pixel are also compared with the class probabilities of pixels from the other two paths of the training algorithm to compute the contrastive losses.}
\label{fig:standard-path}
\end{figure}

\paragraph{Training paths} The training algorithm of AquaCluster contains three training paths: (1) the standard training path, (2) the augmented training path and (3) the augmented shuffled training path. Each of these training paths takes the corresponding image tiles as input and processes them using an encoding sub-model. For each image tile, the encoding sub-model produces a tile containing an encoding for each pixel of the image tile. The tiles of per-pixel encodings are then processed by the prediction sub-model, which outputs a tile containing class probabilities for each pixel. The number of classes that the prediction sub-model outputs probabilities for, denoted as $N_{class}$, is greater than the two ground truth classes we are interested in (water and land). The encoding sub-models in the augmented training path and the augmented shuffled training path share the same parameters. The prediction sub-models in all three training paths share the same parameters. The training paths constitute steps (2), (5) and (8). Figure \ref{fig:standard-path} depicts the standard training path.

\paragraph{Deep clustering losses} The first deep clustering loss is computed from the tiles of class probabilities produced by the standard training path. For each pixel in those tiles, the class with the highest probability is treated as the pixel's label. Using these pseudo-labels, we calculate a weighted cross-entropy loss for each pixel. For each tile, we count the number of pixels that are assigned each class as the pseudo-label and multiply the loss for each class by the inverse of the number of pixels assigned to that class. This process results in classes that are assigned to fewer pixels being given more importance, which encourages the algorithm to use all available classes instead of focusing on only a few. The sum of the weighted cross-entropy loss over each pixel is the first deep clustering loss $L_c$. The same procedure is applied to the augmented tiles of class probabilities produced by the augmented training path to produce the second deep clustering loss $\hat{L}_c$. The deep clustering losses constitute steps (3) and (6).

\paragraph{Negative sampling losses} To include a negative sampling loss we need positive and negative pairs for each data point in our training dataset. Each image tile forms a positive data pair with the corresponding augmented image tile, since they depict the same geographical area. In order to encourage the AquaCluster model to produce similar segmentations for all positive pairs, the sum of the per-pixel absolute difference in class probabilities over each pixel pair is added as a loss to the model. This loss is called the positive spatial consistency loss $L_p$. Similarly, each image tile forms a negative data pair with the corresponding augmented image shuffled tile, since they depict different geographical areas. In order to encourage the AquaCluster model to produce dissimilar segmentations for all negative pairs, the negative sum of the per-pixel absolute difference in class probabilities over each pixel pair is added as a loss to the model. This loss is called the negative spatial consistency loss $L_n$. The negative sampling losses constitute steps (9) and (10).

\paragraph{Total loss} The weighted sum of the four losses mentioned above is the total loss used to train the three sub-models, as shown in Equation \ref{eq:loss}:

\begin{equation}\label{eq:loss}
L=\alpha_c\times(L_c + \hat{L}_c) + \alpha_p\times L_p + \alpha_n\times L_n
\end{equation}

The symbols $\alpha_c$, $\alpha_p$ and $\alpha_n$ denote the weights given to the deep clustering losses, the positive loss and the negative sampling loss respectively. The total loss of the model constitutes step (11).

\paragraph{Encoding sub-models architecture} For the encoding sub-models of AquaCluster we use a modified version of the U-Net model~\cite{ronneberger2015u}, which is designed specifically for the task of semantic segmentation. The U-Net consists of two paths, the contracting path and the expansive path. The contracting path generates an encoding of the input image at a low resolution, while the expansive path upscales the encoding to the original resolution of the image. We modified the architecture by replacing the transposed convolutional layers with simple upsampling layers after encountering the checkerboard patter artifacts that are a known side effect of transposed convolutional layers~\cite{odena2016deconvolution}.

\paragraph{Prediction sub-model architecture} The architecture of our prediction model is much simpler, consisting of a single-layered \gls{CNN}, which transforms the final encoding for each pixel into class probabilities.

\subsection{Inference}
Only the encoding sub-model of the standard training path and the prediction sub-model are used during inference. The input image is processed by the encoding sub-model. The pixel-level encodings produced are then processed by the prediction sub-model, which outputs pixel-level class probabilities. Finally, each pixel is assigned the class for which it has the greatest probability.

\begin{figure}[!hb]
\centering
\includegraphics[scale=0.35]{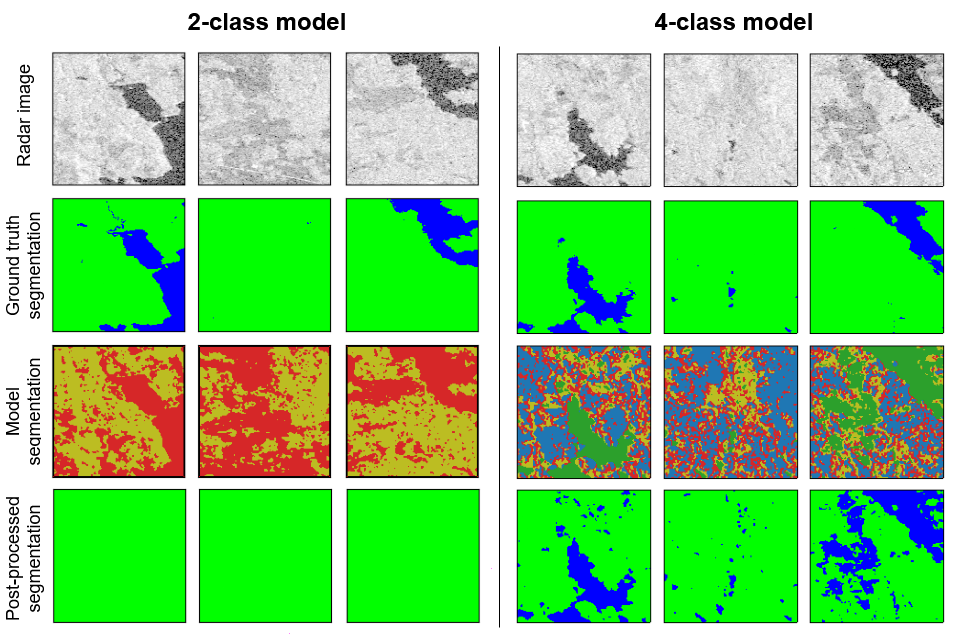} 
\caption{Examples of segmentation when using models with two and four model classes. The first row shows the input radar images, the second row shows the ground truth water/land segmentations, the third row shows the model segmentations and the fourth row shows the land/water segmentations produced by post-processing the model segmentations. As can be seen in the left sub-figure, while the model using two model classes produces a logical segmentation of the input image into two classes, these two classes do not correspond to the water and land ground truth classes. Since both of the model classes end up having a greater overlap with the ground truth land class in the test dataset, they are both assigned to the land class during post-processing. As can be seen in the right sub-figure, the model using four model classes produces segmentations that better separate land from water, although even for this model large parts of land are grouped with water areas.}
\label{fig:k2-4}
\end{figure}

As we mentioned in \ref{sec:back-2}, due to the lack of annotated data to guide AquaCluster, we cannot control what kind of surface each of the model classes corresponds to. Therefore there is not necessarily an one-to-one correspondence between the classes the model outputs and the ground truth classes of the dataset. An illustration of the difference between model classes and ground truth classes, as well as how the number of model classes can affect the performance of AquaCluster, can be seen in Figure \ref{fig:k2-4}. That figure presents two examples, one of a self-supervised model that segments the input images into two classes and one that segments them into four. We can see that while both models produce logical segmentations, neither of them produces segmentations that cleanly separate water from land areas. However the model that segments the images into four classes performs better than the one that segments the images into two, since the higher number of classes allows for more granular segmentations. In our case, we discovered that the number of model classes necessary for the model to be able to create sufficiently detailed segmentations for our task was much higher than the two ground truth classes. Our model outputs probabilities for ten different classes, and each model class is matched to the ground truth class with which it has the greatest overlap in a post-processing step described in Section \ref{sec:post-processing}.

\subsection{Separating model classes into water and land}\label{sec:post-processing}
As we mentioned before, the goal of the model is to generate segmentations that separate water from land, without needing to label each of the classes it produces as one of the two ground truth classes, water or land. However, in order to evaluate the performance of the model, we needed to assign each model class to water or land. We achieved this in a post-processing step, after all model predictions have been made. For each model class, we compared how similar it was to each of the two ground truth classes (water and land) using the \gls{IOU} metric described in Section \ref{sec:eval-metrics} over the whole of the test dataset. Then we assigned each pixel of that model class to the ground truth class it was most similar to for evaluation purposes. 

\subsection{Ensemble models}\label{sec:ensemble}
During our experiments we noticed that the performance of the model varied significantly between experiments. Since the training uses no annotated data, the kinds of classes that the AquaCluster model segments the input images depend a lot on the initialization conditions of the model. This sometimes results in model classes that do not segment water from land areas effectively. Since this problem seems to only affect a small percentage of the models, we decided to create an ensemble of ten models, each voting on whether each pixel of the input image is land or water. This ensemble is created by using the previously described training algorithm to train each model independently of the others on the same dataset. However, since the model classes of each model do not have any correlation with those of the other models, we combine the model predictions after the post-processing step has transformed the model classes into water and land classes. This combination is achieved with per-pixel simple majority voting. For each pixel in the input image we count the number of models that have assigned that pixel to the water or land ground truth class. The ensemble model assigns to that pixel the ground truth class that has the highest count.

\section{Experiments}
\label{sec:experiments}

We conducted experiments to test how well the AquaCluster model can separate water from land from radar images of wetlands. We first trained ten copies of the AquaCluster model on satellite images of a large area of Sweden. Each model was trained on the same dataset, independently of each other, using the training algorithm described in Section~\ref{sec:method}. Afterwards, each of those models was used to produce segmentations for a number of radar images of three Swedish wetlands. Using manually created wand/water segmentations, we converted the model class segmentations of each model into water/land segmentations using the algorithm described in Section~\ref{sec:post-processing}. We then evaluate the land/water segmentations of each model individually, along with the segmentations of the ensemble of those models, created as described in Section~\ref{sec:ensemble}. We compare the segmentations of our models with those obtained using the statistical method Otsu thresholding~\cite{otsu1979threshold} and the optical-based Dynamic World model~\cite{brown2022dynamic}. In this section we present the details about our experiments.
\subsection{Datasets}

\paragraph{Training and validation dataset} The Örebro radar dataset~\cite{pena2023deepaqua} consists of radar satellite images of wetlands in the county of Örebro in Sweden. The images were captured on the date 04/07/2018. These images have a pixel resolution of ten meters. The percentage of water pixels in the dataset is $9.42\%$. After splitting the Örebro area into 639 tiles of $512\times 512$ pixels, $80\%$ of the tiles were randomly selected as the training dataset for our model and the remaining $20\%$ were used as the validation dataset.

\paragraph{Testing dataset} The Swedish Wetlands radar dataset consists of 39 radar satellite images of three different Swedish wetlands (Hjalstaviken, Hornborgarsjon and Svartadalen) on different dates along with manually generated water segmentation masks of those areas. These images were captured on various dates in the years 2018 and 2019 and range in size from $266\times669$ to $1049\times1667$ pixels. In order to avoid images that contain a lot of snow, which can be difficult to differentiate from water in radar images, we did not include any pictures taken during the months of December to March. The dataset has a pixel resolution of ten meters and the percentage of water pixels is $22.27\%$.

\subsection{Evaluation metrics}\label{sec:eval-metrics}
In order to evaluate the performance of the algorithm in the semantic segmentation task, we used several different metrics. Below we provide a short description of each metric.
\paragraph{Accuracy} The most commonly used metric for assessing the performance of a classification model, accuracy describes the percentage of pixels that have been correctly classified and is defined as:
\begin{equation}
Accuracy = \frac{TP + TN}{TP + TN + FP + FN}
\end{equation}
$TP$ represents the number of pixels that have been correctly classified as positive, $TN$ represents the number of pixels that have been correctly classified as negative, $FP$ represents the number of pixels that have been incorrectly classified as positive and $FN$ represents the number of pixels that have been incorrectly classified as negative, For our dataset we consider water the positive class and land the negative class.
\paragraph{Precision} Precision describes the percentage of positive classifications that are correct and is defined as:
\begin{equation}
Precision = \frac{TP}{TP + FP}
\end{equation}
\paragraph{Recall} Recall describes the percentage of ground truth positive pixels that have been correctly classified and is defined as:
\begin{equation}
Recall = \frac{TP}{TP +FN}
\end{equation}
\paragraph{F1-Score} F1-Score is defined as the harmonic mean of the precision and recall metrics:
\begin{equation}
Accuracy = 2\times\frac{Precision\times Recall}{Precision + Recall}
\end{equation}
\paragraph{Intersection over Union}\gls{IOU} is used to compare the similarity between two shapes and is well-suited to semantic segmentation tasks that contain significant class imbalances, which is quite common in land cover datasets~\cite{class_imbalance}. The formula we use for \gls{IOU} is:
\begin{equation}
IoU = \frac{A_{pred} \cap A_{gt} + \epsilon}{A_{pred} \cup A_{gt} + \epsilon}
\end{equation}
where $A_{pred}$ is the area classified as water by our algorithm, $A_{gt}$ is the ground truth water area and $\epsilon$ is a small constant.

\subsection{Baseline Methods}

\paragraph{Otsu thresholding}Otsu thresholding~\cite{otsu1979threshold} is a non-parametric statistical method used to separate a grayscale image into background and foreground by splitting the pixels into two classes based on whether their value is higher or lower than the threshold. The Otsu threshold minimizes the variance of the values in each of the two classes and maximizes the variance of the values between the two classes. Since water surfaces are usually much darker than ground surfaces in radar images, the Otsu thresholding method is used for semantic segmentation of water in satellite images~\cite{pena2023deepaqua}. We used Otsu thresholding as a baseline method that does not use any training data. Following the example of~\cite{pena2023deepaqua}, we applied a Gaussian blur to the images before using the Otsu method to reduce the impact of the noise inherent in radar images.

\paragraph{Dynamic World} The Dynamic World dataset is a land-cover dataset that is produced using optical satellite images and machine learning models~\cite{brown2022dynamic}, with annotations available for all Sentinel-2 satellite images on areas that are cloud-free. Since the Sentinel-2 images are not often available for the same day as the Sentinel-1 images that AquaCluster uses, for each image in our test dataset we looked for available Dynamic World annotations of the test area from 15 days before to 15 days after the radar image. Using this method, and after discarding missing or partial Dynamic World annotations due to clouds, we compiled Dynamic World annotations for 29 of the 39 images in our test dataset.

\section{Results}
\label{sec:results}

\begin{table}[!hb]
  \centering
    \caption{This table shows the performance of the Otsu algorithm, the self-supervised model and the ensemble self-supervised model on the test dataset. The result shown for the non-ensemble AquaCluster model is the mean value obtained over ten experiments.}
  \begin{tabular}{llllll}
  \hline
    \textbf{Model type} & \textbf{Accuracy}& \textbf{Precision}& \textbf{Recall}& \textbf{F1-Score}& \textbf{IoU}\\
    \hline
    Otsu &  $ 0.96$&  $ 0.90$&  $0.89$&  $0.89$&  $0.81$ \\
    Dynamic World &  $0.94$&  $0.87$&  $0.82$&  $0.84$&  $0.73$ \\
    AquaCluster& $0.97$&  $0.88$&  $0.95$&  $0.91$&  $0.85$\\
    \textbf{AquaCluster ensemble}& $\textbf{0.98}$&  $\textbf{0.92}$&  $\textbf{0.96}$&  $\textbf{0.94}$&  $\textbf{0.89}$\\
  \hline
  \end{tabular}
\label{table:big}
\end{table}

In this section, we present the results of our experiments. Table \ref{table:big} compares the results of the four segmentation techniques we tested: (1) the Otsu thresholding algorithm, (2) the Dynamic World annotations, (3) the AquaCluster model and (4) the ensemble AquaCluster model. The AquaCluster result is the mean over ten different experiments, and these same ten models are used to compute the AquaCluster ensemble results. We can see that the optical-based Dynamic World model performs worse than all other methods on all metrics. The Otsu thresholding algorithm outperforms Dynamic World, especially in the Recall and \gls{IOU} metrics. This indicates that Dynamic World miss-classifies water as land significantly more often than Otsu thresholding, and that the shape of the water bodies produced by Otsu thresholding algorithm are closer to the ground truth ones than the shape of the water bodies produced by Dynamic World. The average performance of the AquaCluster models shows a significant improvement in the Recall and \gls{IOU} metrics compared to the Otsu thresholding algorithm, with the other metrics being similar. Finally, the ensemble AquaCluster model outperforms all other methods in all metrics.

\begin{figure*}[!ht]
\centering
\includegraphics[scale=0.26]{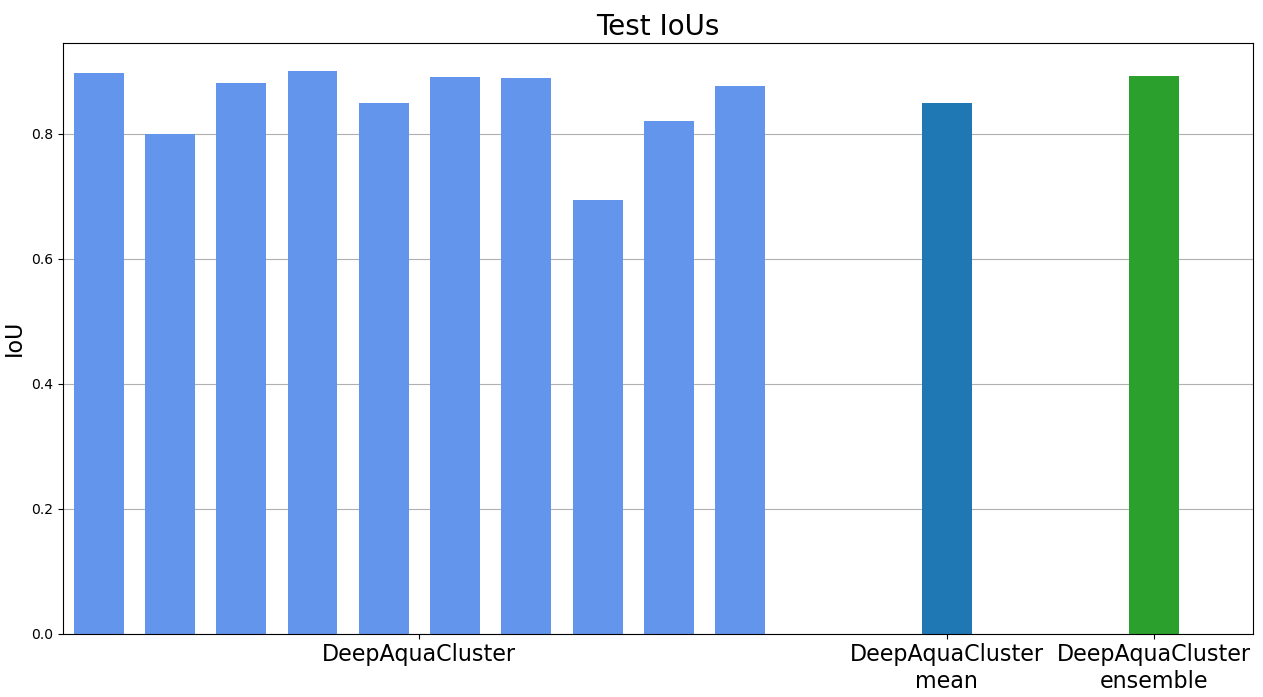} 
\caption{\gls{IOU} test results of the AquaCluster model over ten different experiments, along with the mean \gls{IOU} test result of those experiments and the \gls{IOU} test result of the ensemble of those models. The performance of the single AquaCluster models varies widely between experiments. The ensemble of those models achieves significantly better performance than the mean of the individual models.}
\label{fig:bar-chart}
\end{figure*}

Figure \ref{fig:bar-chart} shows the \gls{IOU} test results of the AquaCluster model over the ten different experiments, along with the mean \gls{IOU} test results of those experiments and the \gls{IOU} test results of the ensemble of these models. As we mentioned in Section~\ref{sec:ensemble}, we observed that the AquaCluster model displays high variance between different experiments. Since the AquaCluster model does not use any annotated data, the characteristics of the different classes that the model segments the input images into are heavily influenced by the random initialization of model weights and the random sequence that the training data are fed to the model. As can be seen in Figure~\ref{fig:k2-4}, some models produce segmentations that include darker land areas along with water areas, which causes those models to misclassify land areas as water. During our experiments we concluded that only a small percentage of the AquaCluster models have this problem. For this reason we implemented the ensemble version of the AquaCluster model as described in Section~\ref{sec:ensemble}. We observe that the ensemble model achieves a significantly higher test \gls{IOU} than the mean test \gls{IOU} of of the individual models.

\begin{figure}[!hbt]
\centering
\includegraphics[scale=0.58]{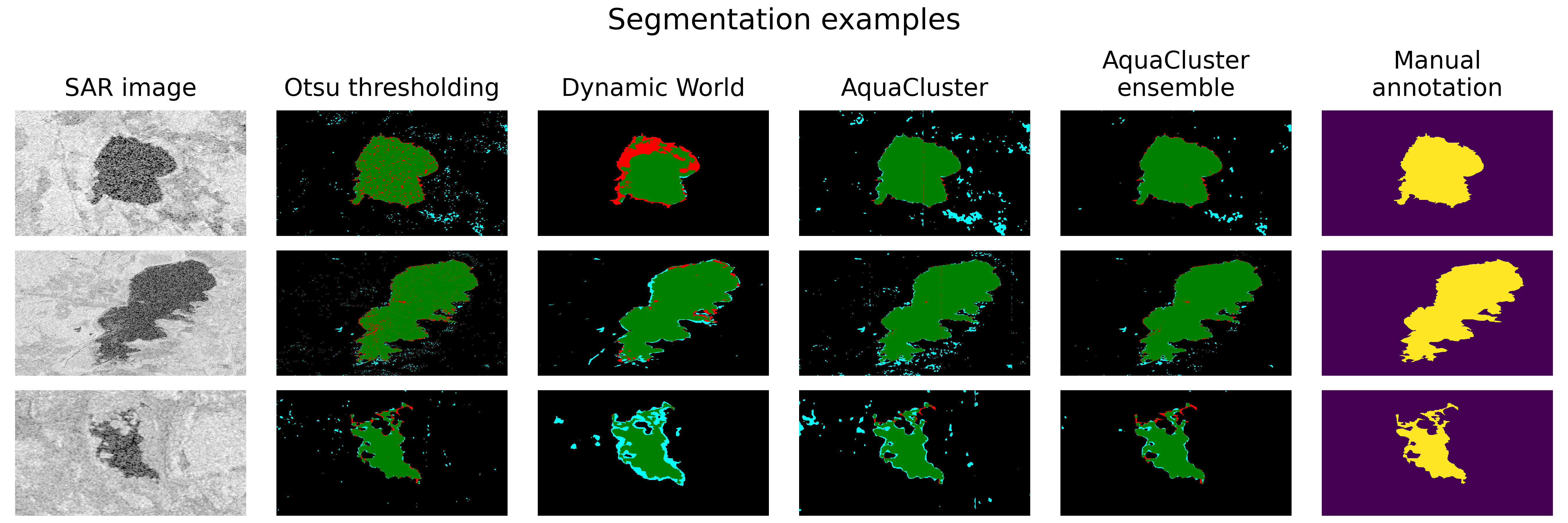} 
\caption{Example segmentations produced by the different segmentation methods. Green denotes true positive segmentations, teal false positive segmentations, black true negative segmentations and red false negative segmentations.}
\label{fig:example-segs}
\end{figure}

Some example segmentations of the different methods for the three test areas can be seen in Figure \ref{fig:example-segs}. We see that the Otsu thresholding algorithm produces very noisy annotations, with numerous very small areas being misclassified. This shows that the Otsu thresholding algorithm has trouble dealing with the noise in the radar images. The Dynamic World model produces less noisy annotations, but it has poor performance at the boarder regions between water and land areas. The single AquaCluster model produces segmentations without the problems of either of the previous methods, but also classifies some of the slightly darker soil areas on the radar images as water. Finally, the segmentations produced by the ensemble of AquaCluster models produces segmentations similar to those of the single AquaCluster model. However, the patches of land that have been misclassified as water are significantly reduced. From these examples we can see that, apart from the differences in general performance, each method has difficulties with different parts of the data.

\section{Conclusion}
\label{sec:conclusion}

In this paper we presented AquaCluster, a \gls{CNN}-based machine learning model that segments radar satellite images into areas that separate open and vegetated water from land. We used a combination of two self-supervised training methods, deep clustering and negative sampling, to train the AquaCluster model using only radar satellite images and no manual annotations. Moreover we implemented an ensemble version of the model to reduce variance and improve performance. The model was tested on radar satellite images of three different Swedish wetlands taken at different dates over two years. The ensemble AquaCluster model shows good performance on the test set and outperforms the statistical radar-based method Otsu thresholding and the optical-based machine learning model Dynamic World. The performance of our model proves that it is possible to train a machine learning model to separate water from land in satellite radar images without the use of any annotations. The AquaCluster training algorithm eliminates the need for annotated data, which are time-consuming and expensive to produce, and clear optical images, which are often unavailable. Thus the AquaCluster model can be easily retrained on new radar satellite images in order to adapt the model to changes such as different climates or changes in satellite sensors.

\section*{Acknowledgments}

This work was supported by Digital Futures . We gratefully acknowledge the Bolin Centre for Climate Research for providing us with a GPU server and the National Academic Infrastructure for Super computing in Sweden for giving us access to their high computing cluster. We thank Ezio Cristofoli and Johanna Hansen for their ideas and for annotating data.

\section*{Data availability}

All the source code, testing dataset, and pre-trained models can be found in our online open-source repository at https://github.com/melqkiades/deep-wetlands. Please cite this article when using the AquaCluster model.
 \bibliographystyle{elsarticle-num} 
 \bibliography{egbib}






\end{document}